\documentclass{article}

\usepackage{arxiv}

\usepackage[utf8]{inputenc} 
\usepackage[T1]{fontenc}    
\usepackage{hyperref}       
\usepackage{url}            
\usepackage{booktabs}       
\usepackage{amsfonts}       
\usepackage{nicefrac}       
\usepackage{microtype}      
\usepackage{lipsum}		
\usepackage{graphicx}
\usepackage{pifont}
\usepackage[sort,numbers]{natbib}
\usepackage{doi}

\usepackage{outlines}
\usepackage[color=gray]{todonotes}
\usepackage{xfp}
\usepackage{numprint}
\usepackage[inline]{enumitem}
\usepackage[binary-units=true]{siunitx}
\usepackage[]{booktabs}
\usepackage[version=3]{mhchem}

\usepackage[figuresright]{rotating}
\usepackage[nolist]{acronym}
\usepackage{xcolor}

\newcommand*\OK{\ding{51}}

\title{Towards a Deep Learning-based Online Quality Prediction System for Welding Processes}

\author{
\href{https://orcid.org/0000-0003-4046-2990}{\includegraphics[scale=0.06]{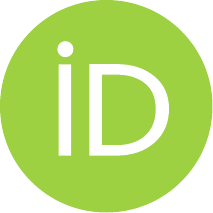}\hspace{1mm}Yannik Hahn}\thanks{Corresponding author. {\it E-mail address:} yhahn@uni-wuppertal.de}, 
\href{https://orcid.org/0000-0001-7372-6898}{\includegraphics[scale=0.06]{orcid.pdf}\hspace{1mm}Robert Maack}, 
\href{https://orcid.org/0000-0003-0080-6285}{\includegraphics[scale=0.06]{orcid.pdf}\hspace{1mm}Hasan Tercan}, 
\href{https://orcid.org/0000-0002-1969-559X}{\includegraphics[scale=0.06]{orcid.pdf}\hspace{1mm}Tobias Meisen} \\
Institute for Technologies and Management of Digital Transformation (TMDT)\\ 
University of Wuppertal \\
Rainer-Gruenter-Straße 21, 42119 Wuppertal, Germany \\ 
\And 
Marion Purrio, 
Guido Buchholz, 
Matthias Angerhausen \\
\textit{Forschungs- und Entwicklungsgesellschaft Fügetechnik GmbH (FEF)}\\
Driescher Gässchen 5, 52062 Aachen, Germany
}

\renewcommand{\shorttitle}{Towards a Deep Learning-based Online Quality Prediction System for Welding Processes}

\hypersetup{
pdftitle={A template for the arxiv style},
pdfsubject={q-bio.NC, q-bio.QM},
pdfauthor={David S.~Hippocampus, Elias D.~Striatum},
pdfkeywords={First keyword, Second keyword, More},
}

\begin{document}
\maketitle

\begin{abstract}

The digitization of manufacturing processes enables promising applications for machine learning-assisted quality assurance. A widely used manufacturing process that can strongly benefit from data-driven solutions is \ac{GMAW}. The welding process is characterized by complex cause-effect relationships between material properties, process conditions and weld quality. In non-laboratory environments with frequently changing process parameters, accurate determination of weld quality by destructive testing is economically unfeasible. Deep learning offers the potential to identify the relationships in available process data and predict the weld quality from process observations. In this paper, we present a concept for a deep learning based predictive quality system in \ac{GMAW}. At its core, the concept involves a pipeline consisting of four major phases: collection and management of multi-sensor data (e.g. current and voltage), real-time processing and feature engineering of the time series data by means of autoencoders, training and deployment of suitable recurrent deep learning models for quality predictions, and model evolutions under changing process conditions using continual learning. The concept provides the foundation for future research activities in which we will realize an online predictive quality system for running production.
\end{abstract}


\begin{acronym}
  \acro{AL}[AL]{arc length}
  \acro{CNN}[CNN]{convolutional neural network}
  \acro{DIN}[DIN]{German Institute for Standardisation Registered Association}
  \acro{DNP}[DNP]{distance from nozzle to plate}
  \acro{EWC}[EWC]{elastic weight consolidation}
  \acro{GFR}[GFR]{gas flow rate}
  \acro{GMAW}[GMAW]{gas metal arc welding}
  \acro{HIVE-COTE}[HIVE-COTE]{}
  \acro{LSTM}[LSTM]{long short-term memory}
  \acro{MAG}[MAG]{metal active gas}
  \acro{MAS}[MAS]{memory-aware synapses}
  \acro{MIG}[MIG]{metal inert gas}
  \acro{RNN}[RNN]{recurent neural network}
  \acro{VPRS}[VPRS]{variable precision rough set}
  \acro{WFR}[WFR]{wire feed rate}
  \acro{WS}[WS]{welding speed}
\end{acronym}

\section{Introduction}\label{ch:introduction}
Joining technologies represent a key position in modern manufacturing. Especially fusion welding processes such as \ac{GMAW} are widely used in fields such as automobile, naval, and aeronautic industries~\cite{Bestard.2018}. \ac{GMAW} relies on electrical energy and its transformation into heat to join metallic materials and is considered one of the most efficient welding techniques for automated welding applications due to its high product yield, high reliability, and good automation capacity. Traditionally, the identification of appropriate setting parameters of the \ac{GMAW} process is a challenging task that highly depends on the operator's long-time experience and expertise. Such process parameters are, for example, the welding speed, electrical cycle time, and material composition of the workpiece and welding wire. To assess the quality of the welding process, the operator primarily observes characteristic effects of the process such as the arc that builds up in the protective gas between the welding wire and the workpiece or the sound of the process that results from cyclical ejections of the liquified welding wire. The assessment of the weld quality involves the inspection of the materialistic composition and geometrical properties of the welding seam. One of the most reliable methods is to cut the welding seam and the welded workpiece transversally and inspect the resulting micro-sections. However, the costs for such destructive inspection methods are disproportionate to the expected revenue, especially for small batch sizes. Non-invasive methods like material scanning with x-ray resolve the problem of high material costs but are expensive, require special precautions to maintain a safe working environment, and are also challenging to configure for a specific welding setup and material composition.

In the present era of Industry 4.0 and the digitization of welding processes, new possibilities are emerging to tackle these challenges. With the ability to collect process data from electrical measurements and sensors and to bring them together with quality measurements, there is great potential for predictive quality solutions \cite{Tercan.2022b}. Predictive quality refers to the use of deep learning to predict the weld quality based on welding process data. Such deep learning models can capture the complex cause-effect relationships between material properties, process conditions, and weld quality. On the one hand, utilizing quality predictions makes it possible to carry out quality assessments that were previously infeasible. And on the other hand, this allows for quality-enhancing measures to be taken during the manufacturing process, such as ad-hoc adjustments of process parameters. In this paper, we present a concept for a deep learning-based online quality prediction system that is to be used in the running \ac{GMAW} process on available measurement data. For this purpose, we address three key aspects. First, we identify quality-influencing process parameters and describe the characteristics of relevant sensor data (e.g. current and voltage). Then, we discuss suitable data processing methods and deep learning models for quality prediction. Finally, we address the question of how the deep learning models can be efficiently trained in the ongoing welding process. Our concept includes the application of continual learning methods to train dynamically evolving deep learning models over sudden or gradual process changes.

The paper is organized as follows: Section \ref{ch:use_case} introduces the basics of the \ac{GMAW} process and its data characteristics. Section \ref{ch:related_work} presents related research works dealing with predictive quality in welding processes and continual learning. Section \ref{ch:concept} describes the proposed concept for an online predictive quality system. Finally, Section \ref{sec:conclusion} briefly concludes the paper and gives an outlook on future research.

\section{Welding Process and Sensory Data}\label{ch:use_case}
\subsection{Gas Metal Arc Welding}
Arc welding is a specific type of fusion welding that is used to join metallic objects~\cite{Weman.2003}. As such, it belongs to the subcategory of joining methods in manufacturing engineering and is well-defined by the \ac{DIN} as DIN 8593-6. In contrast to other welding techniques like gas or laser welding, arc welding utilizes electrical energy to create the heat necessary to melt welding wire and workpiece. In this paper, we focus on \ac{GMAW} which combines a continuously fed welding wire as the electrode and shielding gas, which protects the material against the influences of the atmosphere. 

\begin{figure}[t]
  \centering \includegraphics[width=0.5\linewidth]{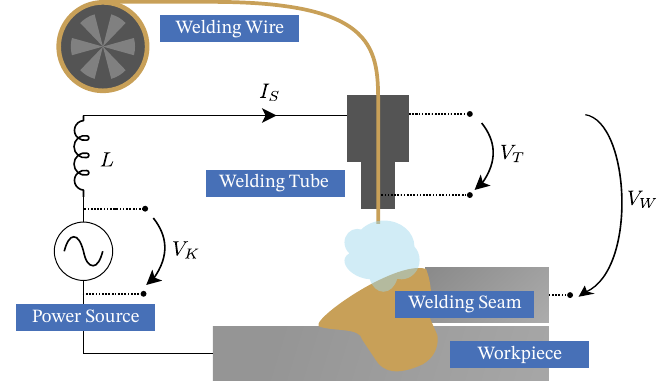}
  \caption{Schematic representation of a welding process with description of the electrical properties as a circuit diagram.}
  \label{fig:illustration_electrical_properties_welding_process}
\end{figure}

The procedurally and physically most important part of the welding setup is situated between the tip of the welding wire and the workpiece. When high voltage is applied between them, an electric arc forms as a result of the ionization of the injected protective gas. When the current density in the welding wire is sufficiently high, it begins to heat up and liquify. This consequently creates metal droplets that fall on the surface of the workpiece creating a melting pool. With the progression of the welding, the melting pool dissipates its heat into the surrounding material, solidifies, and creates a welding seam that fuses the previously separated metal sheets. Figure \ref{fig:illustration_electrical_properties_welding_process} shows the electrical circuit of a typical \ac{GMAW}-setup.

The energy source is configured such that the terminal voltage $V_{K}$ drives a current with certain characteristics that influence the behavior of the welding process. The total voltage drop can be approximated with the \textit{welding voltage} $V_{W}$ over resistance at the sliding contact between contact tube and wire, wire resistance inside the ontact tube and resistance of the arc. The voltage drop along the conductors from and to the current source, along the part of the wire before the sliding contact and in the workpiece itself are neglectable. The control loop mainly incorporates observations of the welding voltage and therefore is our main focus point. Before using a new welding setup, the control loop estimates its physical properties. Modern setups achieve this by initially measuring the inductivity and resistance. Especially the resistance at the sliding contact is unavoidably subject to wear which changes its physical properties over time.

\subsection{Quality Metrics}\label{sec:quality_metrics}
The quality levels of fusion-welded joints of various metals and their alloys are well-defined in \ac{DIN} EN ISO 5817:2014 which considers typical indicators for degraded quality such as nonstandard weld geomtries, weld penetration, internal fusion, cracks, porosities and gaps.
Besides thermally induced geometrical distortion and unintentional changes in local material properties due to welding heat impact, the most relevant product quality indicator is the geometry of the weld seam and the resulting internal fusion geometry. The tolerable sizes and dimensions of these imperfection are dependent upon design-based functional loads and the thicknesses of the parts to be welded. The resulting weld quality is essentially classified binarily with OK or NOK (not OK), depending upon the size and relations of present imperfections.\
In general, the product quality is determined offline after the joining has been conducted. This mostly involves visual inspection by domain experts. For our research, we incorporate photos taken from the top and bottom of the joined workpiece as well as photos of microsections to access the internal geometrical structure of the weld. A more fine-grained differentiation of the weld quality, although potentially possible, is currently not used in practice. However, it is potentially worthwhile to expand the number of discriminating categories to enhance the expressiveness of our deep learning quality prediction models.

\subsection{Sensory Data of Current and Voltage}
\label{sec:observed_process_parameters}
\begin{figure}[t]
  \centering \includegraphics[width=0.5\linewidth]{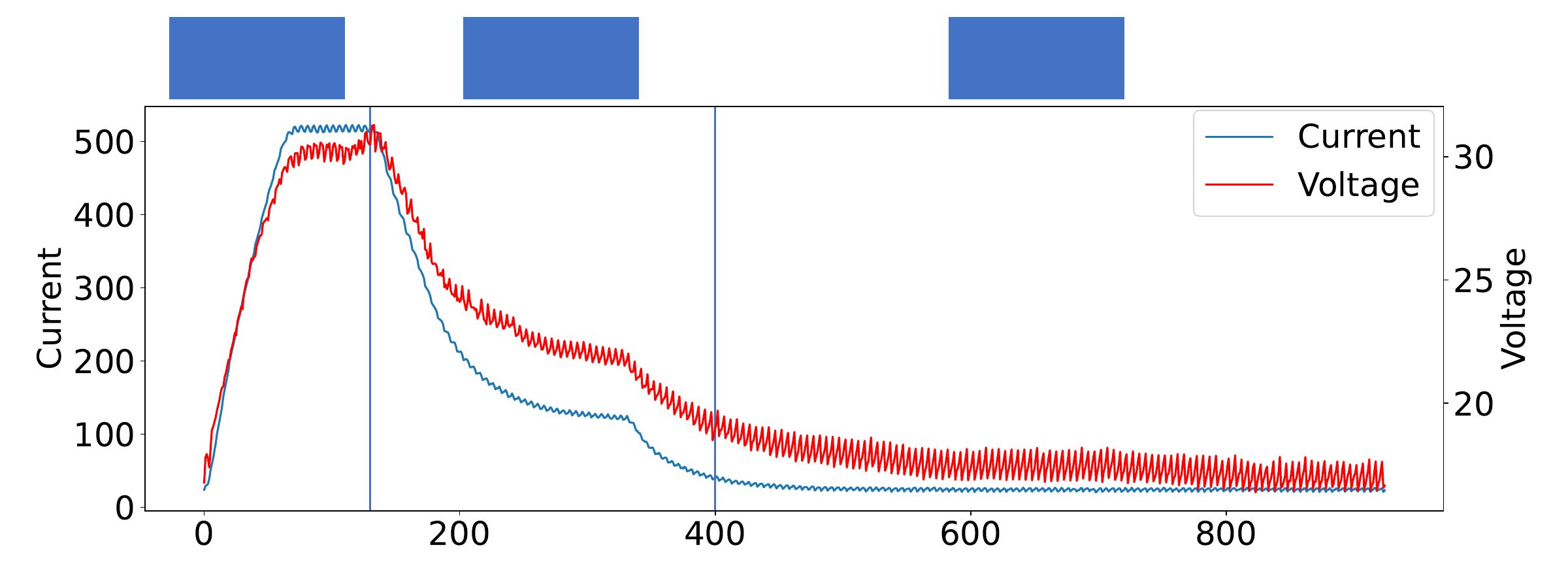}
  \caption{Example of one cycle split into the described three distinct phases.}
  \label{fig:one_cycle}
\end{figure}

As motivated in Section \ref{ch:introduction}, our focus lies in the characterization of the welding process behavior and inference of the final weld quality by observing process sensory data. Among all sensory data, the current $I_{S}$ and the welding voltage $V_{W}$ are considered the most influential for the estimation of the weld quality given that electrical power is responsible for liquification of the welding wire. All currents and voltages can be synchronously sampled with the sampling frequency of \SI{100}{\kilo\hertz} with a maximum error of \SI{0.5}{\percent}. The sequence $x_t \in X=\{x_1, .., x_n\}$, with $x_t \in \mathbb{R}^{n \times 2}$ represent the corresponding multivariate time series. $X$ can be subdivided into cycles $c_i \in C = \{c_1, ..,c_m\}$, where $C\subset X$ and $m$ is the number of cycles in one welding process. Figure \ref{fig:one_cycle} illustrates an exemplary cycle each for the current and the voltage. Each cycle can be split up into three non-overlapping phases $c_i=p^1_i, p^2_i, p^3_i$, each of which has a distinct role for the build-up of the droplet and subsequent detachment into the melting bed:
\begin{enumerate}[label=Phase \arabic*, start=1,leftmargin=*,resume]
  \item \textbf{Pulse phase}: The current source increases the current flow which heats up and liquefies the welding wire. A steep ramp-up of current is required such that a sufficient amount of energy can be transferred quickly to form a droplet with distinct dimensions. The phase ends when the current reaches a plateau.
  \item \textbf{Droplet detachment phase}: As soon as the droplet reaches a certain size, it detaches from the welding wire and is ejected toward the melt pool. Ideally, the induction of energy is reduced such that the liquefication of the welding wire slows down to reduce the risk of a droplet being deformed excessively in the longitudinal direction, potentially leading to an undesirable short circuit.
  \item \textbf{Base current phase}: The current flow is further decreased, reaching the bottom \SI{5}{\percent} of the maximum power in the current cycle. This ensures the next cycle starts in an energetically well-defined state and the formation of a new droplet is initiated.
\end{enumerate}

In a theoretically ideal setup, a single droplet builds up and is ejected into the melting pool in each cycle. In reality, however, the physical behavior is not always exactly as described. For example, a droplet may not detach from the wire as expected and will continue to build up mass throughout one or more following cycles. Another common case is that the droplet is not falling into the melt pool but makes contact with the workpiece. The behavior in a single cycle is considered to be stochastic and has no significant effect on the overall quality of the weld. However, sequences across multiple cycles as a whole provide valuable information and thusly relevant data to estimate the quality of the weld. That in turn has implications on the design of the pre-processing pipeline and the deep learning model for the quality prediction (see Section~\ref{ch:concept}). 

\section{Related Work}\label{ch:related_work}
\subsection{Quality Prediction in Welding Processes}
The \ac{GMAW} process is influenced by multiple parameters that impact the quality of the welding process. In the current state of research on data-driven quality prediction in welding, images and sensor data are the most commonly used input variables for the models. Table~\ref{tab:sensor_types} lists the publications primarily using sensor data for model training~\cite{Zhang.2022, Daniyan.2019, Gyasi.2019b, Singh.2021, Li.2008, Sartipizadeh.2018, WangXuanyin.2010, Nagarajan.2019, Kesse.2020}. It shows that current  (around 78\% of publications) and voltage (around 55\%) are among the most used input types. In contrast to our approach, which aims to include features over multiple cycles to capture the stochastic behavior of the processes, all publications used sensor inputs as aggregated values over a welding process.

A welding quality prediction task can be defined as a classification of categorical quality metrics or a regression of numerical metrics. Those publications that perform classification predict quality metrics such as weld bead width~\cite{Kesse.2020, Li.2008}, or weld distortion and hardness~\cite{Daniyan.2019}. On the contrary, the publications performing regression directly predict bead geometry values~\cite{Singh.2021, Gyasi.2019b, Daniyan.2019, Nagarajan.2019}. Both mathematical and deep learning models are utilized in the context of predictive quality for arc welding. The mathematical methods include the Taguchi model~\cite{Daniyan.2019}, adaptive iterative model~\cite{WangXuanyin.2010}, and bayesian networks~\cite{Nagarajan.2019}. These models require carefully handcrafted features and models. On the contrary, deep learning models directly extract and recognize patterns in the data. Deep learning has recently shown high potential on sensory data in the industrial context for quality control~\cite{Meyes.2021,Maack.2022} and predictive maintenance~\cite{Maack.2021}. \citet{Bestard.2018} reviewed the development of arc welding and identified artificial neural networks as one of the most used methods in a summary of the last 50 years of arc welding. In their literature review on adaptive robotic welding, \citet{Zhang.2021} identified that publications predicting the welding performance used neural networks. Likewise, \citet{Zhang.2022} used a two-layer neural network in a setting with aggregated values per welding process. \citet{Gyasi.2019b} proposed a neural network with three hidden layers to predict the wire feed rate. \citet{Li.2008} compared a neural network with three hidden layers to a \ac{VPRS}. On the contrary, \citet{Kesse.2020} generated a dataset using fuzzy logic to train a fuzzy-driven neural network with six hidden layers.

The aforementioned neural networks, which are small compared to state-of-the-art deep learning models, were sufficient for strongly aggregated values over one welding process. However, these aggregated values do not represent the underlying data's stochastic trends and physical interdependencies. Therefore, we will focus on capturing those characteristics by utilizing deeper and more advanced deep learning models, like \acp{LSTM}, which are needed due to the higher complexity.

\begin{table}
    \small
    \centering
    \begin{tabular}{llllllll}
        Paper &
        \acs{WS} & Current & \acs{WFR} & \acs{GFR} & \acs{AL} & \acs{DNP} & Voltage \\
        \hline
        \cite{Zhang.2022} & \OK & \OK & \OK &  &  &  & \OK \\
        \cite{Daniyan.2019} & \OK & \OK &  &  & \OK &  & \OK \\
        \cite{Gyasi.2019b} & \OK & \OK & \OK &  &  &  &  \\
        \cite{Singh.2021} & \OK &  & \OK &  &  & \OK & \OK \\
        \cite{Li.2008} & \OK & \OK &  &  & \OK &  &  \\
        \cite{Sartipizadeh.2018} &  &  & \OK &  &  &  & \OK \\
        \cite{WangXuanyin.2010} &  & \OK & \OK &  &  &  &  \\
        \cite{Nagarajan.2019} & \OK & \OK & \OK & \OK & \OK &  &  \\
        \cite{Kesse.2020} & \OK & \OK &  &  &  &  & \OK \\
        \hline
    \end{tabular}
    \caption{Summary of the extracted sensors types which are used as input variables in the publications: \acf{WS}, current, voltage, \acf{AL}, \acf{GFR}, \acf{WFR}, \acf{DNP}}\label{tab:sensor_types}
\end{table}
\subsection{Continual Learning}
The described works in predictive quality are mainly based on the assumption that the training data basis is representative for the given problem. However, this assumption is not valid in processes such as GMAW, as they are subject to continuous changes (e.g. changes in main setting parameters or material properties). Process changes mean that previously trained learning models no longer work sufficiently well, which can be caused by out-of-distribution predictions. Consequently, new training data has to be generated at a great cost and time. Continual learning is a paradigm that tackles this problem and deals with training neural networks over time in such a way that new knowledge can be acquired for new tasks and catastrophic forgetting from previously trained tasks is minimized~\cite{Parisi.2019}. On the one hand, there exist rehearsal methods that use a fixed-sized memory to store data samples from previously trained tasks. These samples are then later revisited during the training of new tasks. Another class of methods are regularization strategies, which reduce catastrophic forgetting by restricting network parameter updates while training on new tasks~\cite{Kirkpatrick.2017, Aljundi.2018}. Concerning manufacturing, few works exist yet on the use of continual learning methods. In~\cite{Tercan.2022}, neural networks are continually trained using regularization for predictive quality tasks in injection molding. In~\cite{Maschler.2020}, the authors use regularization for fault prediction of turbofan engines based on \ac{LSTM} networks which shows that real-world predictive quality approaches for continual learning utilizing sensor data exist, suggesting transferability to our problem.
\section{Concept for Online Quality Prediction in Welding}\label{ch:concept}
\begin{figure*}[htbp!]
  \centering \includegraphics[width=\linewidth]{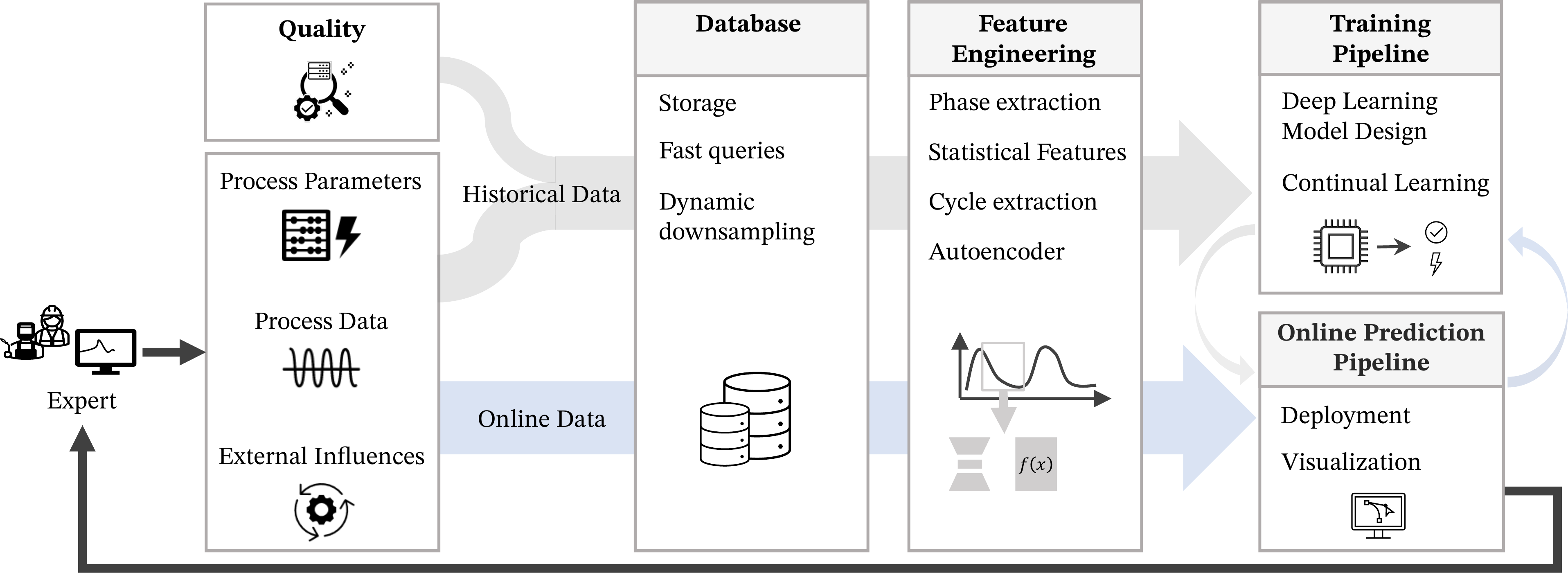}
  \caption{Illustration of proposed predictive quality concept for welding, combining the collection of welding data, training on historical data, and online prediction.}
  \label{fig:full_process_concept}
\end{figure*}
Our concept builds upon the described \ac{GMAW} process and the measured process data. The concept involves a self-contained processing and prediction pipeline. We distinguish two different data pathways given that, on one hand, we need to be able to design, train and adapt a deep learning architecture that has fundamental quality prediction capabilities, and on the other, an online prediction system that makes effective use of the trained deep learning model that resembles a feedback loop between prediction system and the machine operator. Figure~\ref{fig:full_process_concept} illustrates that pipeline. In the following, its components are described in detail. First, the (pre-)processing steps, including the Database (\ref{sec:data_management}) and Feature Engineering (\ref{ch:feature_engineering}), and then Model Training (\ref{ch:deep_learning_model}) and Online Prediction (\ref{ch:continual_leanring}) in which the deep learning model and continual learning approaches are implemented.

\subsection{Data Management and Visualization}\label{sec:data_management}
The storage of time-encoded sensory data from continuous process observations is a major challenge that requires precautions in the selection of an adequate storage solution. On the one hand, large storage capacity has to be provided to store the sensory data acquired with a high sampling rate. Given that the actually necessary sampling rate for the training of a deep learning model is yet unknown, it is advisable to store as much information as possible. On the other hand, data access times are not crucial during the training phase but more so for the inference phase, in which online data has to be immediately accessible with reasonable latencies. Consequently, we utilize TimescaleDB, a PostgreSQL-based database specifically designed to optimize queries on time series data~\cite{timescaledb}, which fulfills our requirements. Data in non-sequential data formats like categorical quality data and numerical process parameters can be stored in any relational database. Given that the amount of samples of that data modality is relatively small in comparison to those of sensory time-series data, they are neglectable regarding runtime efficiency.

\subsection{Feature Engineering}\label{ch:feature_engineering}
\begin{figure}[t]
  \centering
  \includegraphics[width=0.5\linewidth]{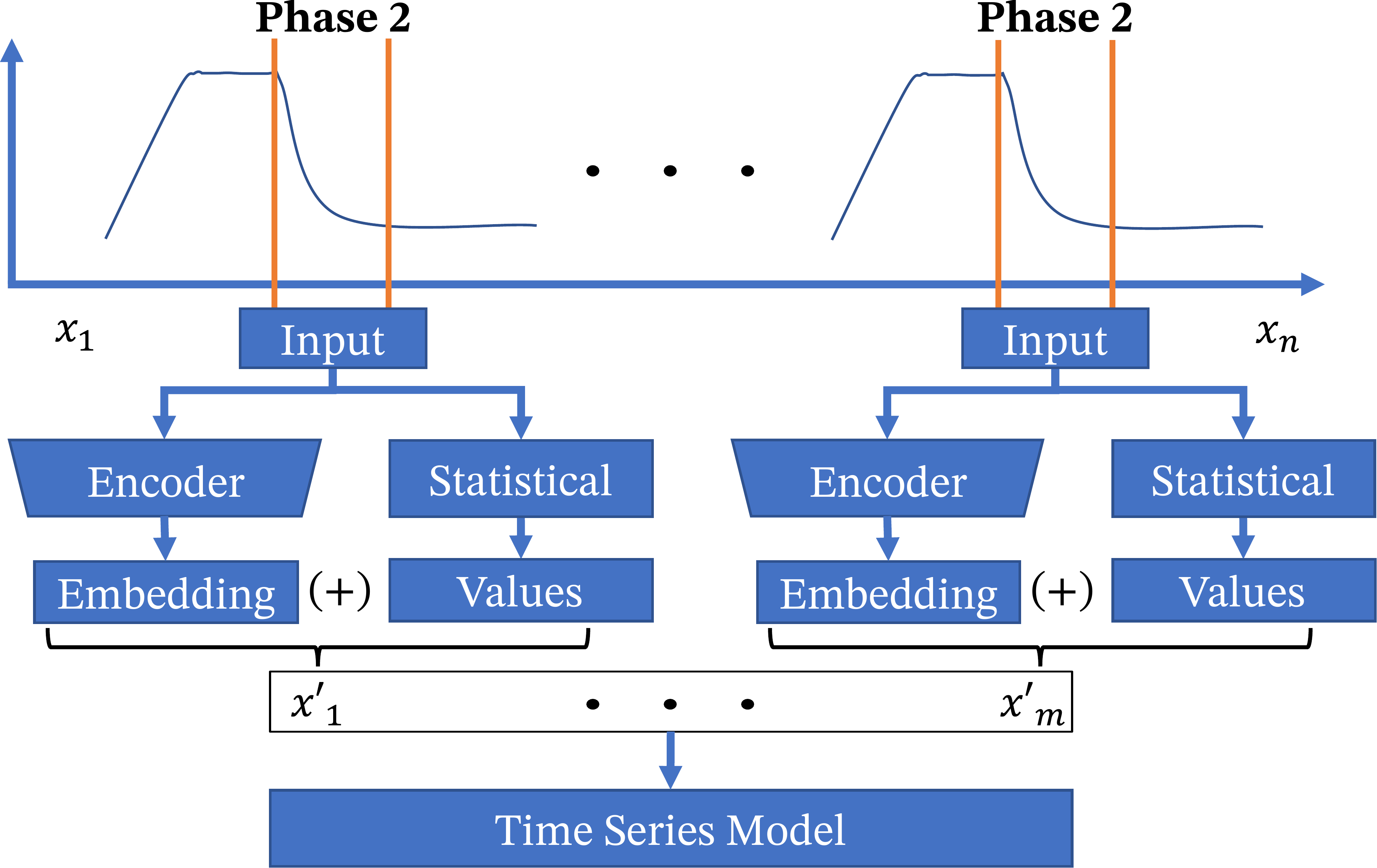}
  \caption{Visualisation of the feature extraction combining an autoencoder approach with statistical values. The extracted features serve as inputs for the time series model.} \label{fig:feature_engineering}
\end{figure}

Furthermore, the data needs to be preprocessed and characteristic features need to be extracted. Figure~\ref{fig:feature_engineering} shows an illustration of the extraction of important features from the three phases of a droplet build-up and detachment cycle. It presents how we plan to use these features with a time series model to predict the weld quality. The pipeline starts by extracting the different cycles $c_i$. Then each cycle is split into the three phases $p^{u}_i$. We will primarily focus on the second phase $p^2_i$ because this phase includes the detachment of the droplet which has a high impact on the resulting quality. To capture the information of thermo-physical influences from the metal welded before, multiple phases $p^2_i$ from coherent cycles need to be used as input for the next processing blocks. Each phase $p^2_i$ is compressed using statistical measurements like minimum, maximum, mean, trend, frequency, etc. In addition, an embedding is trained by a deep learning model to give a compressed representation of this one phase. For this task, an autoencoder is used to compress the information of the data through a bottleneck by predicting the initial input in an unsupervised manner~\cite{bank2020autoencoders}. After training the autoencoder, the resulting encoder part of the autoencoder is utilized to compress the input sequence $p^2_i$ into an embedding. The statistical values and the embeddings are then concatenated to a feature vector $x^{\prime}_{t^{\prime}}$ for each cycle. Next, these feature vector $x^{\prime}_{t^{\prime}}$ gets combined with adjacent feature vectors to a sequence $X^{\prime}=x^{\prime}_{1}, .., x^{\prime}_{m}$ where $m$ denotes the number of cycles. Then the resulting sequence $X^{\prime}$ is used as input for a time series deep learning model.

\subsection{Data-processing and Deep Learning}\label{ch:deep_learning_model}
The online prediction pipeline uses the same feature preprocessing and feature extraction pipeline as the training pipeline. A deep learning model is trained on historical process data and then deployed for online quality predictions. The predictions are fed back to the process expert via a real-time visualization. Due to the high sampling rate of a welding process and the fine-grained underlying characteristics of the data, a preprocessing of the data is needed. Central challenges in processing the data are the long sequences and the long-term inter-dependencies a model needs to capture. Long sequences result in high computational costs of time series models, which grow with the length of the sequence. Therefore, a strong compression that keeps the quality-related information of the input data is needed.

It has been shown that multiple neural network architectures were used to predict weld quality. Our concept involves time series models, which are deep learning models that leverage the characteristics of time series data specifically. Time series models have been extensively researched for classification tasks in real-world environments~\cite{ismail2019deep}. Long short-term memory \ac{LSTM} models are known to work well on sequence processing tasks because they have an improved remembering capacity compared to standard recurrent cells \cite{yu2019review}. Consequently, we will use a \ac{LSTM} to encode the sequence $x^{\prime}_{1}, ..,x^{\prime}_{m}$ which is the result of the previously defined preprocessing pipeline. We will add an additional linear layer after the \ac{LSTM} which is needed to make a classification. However, due to the modular way of constructing our concept, the time series model can be switched to a better-performing time series model in future applications if needed.

\subsection{Continual Learning}\label{ch:continual_leanring}
Due to the dynamics of the welding process and changes in boundary conditions and process parameters in the running production, it must be possible to continuously update the deployed time series model during its usage phase. This means that, as soon as significant process changes occur, the model needs to be trained on new representative process observations. As described in Section \ref{ch:related_work}, continual learning methods and especially regularization strategies are suitable in this context. Our solution concept involves the use of regularization strategies such as \ac{EWC} ~\cite{Kirkpatrick.2017} or \ac{MAS} ~\cite{Aljundi.2018}. They ensure that the underlying neural network does not forget previous knowledge when trained on new process data. We mainly consider process changes that occur frequently in the arc welding process. These are the change of quality-determining process parameters (e.g. wire feed rate or welding speed) and the change of welding equipment or hardware components (e.g. power source or welding head). Each of these changes triggers, identified by out-of-distribution process parameters, a model update that ensures that the model can be used for process observations. The model update involves two main steps. First, a new representative data set is created by collecting new experimental observations from the welding process. Then, the model is re-trained on the data using \ac{EWC} or \ac{MAS}. In this process, only the most recently collected data needs to be used for training. Finally, the updated model is deployed to be used for quality predictions.

\section{Conclusion and Outlook}\label{sec:conclusion}
In this paper, we presented a concept for the application of deep learning-based predictive quality in \ac{GMAW} -- one of the most popular joining methods for metal components. We argue that current approaches do not exhaustively leverage the observable data from the welding process' current and voltage measurements, although available in many instances. Instead, we motivate the utilization of state-of-the-art deep learning models that extract characteristic features from the underlying time-series data and continual learning methods to efficiently update the models across welding process variations. In our future work, we will take this concept as a foundation to develop, implement and evaluate a quality prediction system in a real welding process. In addition, we will contribute to the research community by publishing the full set of sensory and quality data for public use~\cite{hahn_yannik_2022_7023254}.

\bibliographystyle{unsrtnat}
\bibliography{references}  

\end{document}